\documentclass[journal]{IEEEtran}
\usepackage{algorithm}
\usepackage{algorithmic}
\usepackage{graphicx}
\usepackage{caption,subcaption}
\usepackage{float}
\usepackage{amsmath}
\usepackage{amssymb}
\usepackage{mathrsfs}
\usepackage{booktabs}
\usepackage{bbm}
\usepackage{multirow}

\usepackage{pifont}
\newcommand{\XSolidBrush}{\ding{55}} 
\usepackage{amssymb}
\newcommand{\CheckmarkBold}{\ding{51}} 

\usepackage[pagebackref,breaklinks,colorlinks]{hyperref}

\ifCLASSINFOpdf
\else
\fi
\UseRawInputEncoding

\begin{document}
%
\title{Exploring Stronger Transformer Representation Learning for Occluded Person Re-Identification}
%
%
%

\author{Zhangjian ~Ji,~\IEEEmembership{Member,~IEEE,}
        Donglin ~Cheng and Kai ~Feng
\thanks{Zhangjian Ji, Donglin Cheng and Kai Feng are with the Key Laboratory of Computational Intelligence
and Chinese Information Processing of Ministry of Education, School of Computer and Information Technology, Shanxi University, Taiyuan 030006, China E-mail: \{jizhangjian@sxu.edu.cn\}}
}

%
%

\markboth{Journal of \LaTeX\ Class Files,~Vol.~14, No.~8, November~2022}%
{Shell \MakeLowercase{\textit{et al.}}: Bare Demo of IEEEtran.cls for IEEE Journals}
%



\maketitle

\begin{abstract}
 Due to some complex factors (e.g., occlusion, pose variation and diverse camera perspectives), extracting stronger feature representation in person re-identification remains a challenging task. In this paper, we proposed a novel self-supervision and supervision combining transformer-based person re-identification framework, namely SSSC-TransReID. Different from the general transformer-based person re-identification models, we designed a self-supervised contrastive learning branch, which can enhance the feature representation for person re-identification without negative samples or additional pre-training. In order to train the contrastive learning branch, we also proposed a novel random rectangle mask strategy to simulate the occlusion in real scenes, so as to enhance the feature representation for occlusion. Finally, we utilized the joint-training loss function to integrate the advantages of supervised learning with ID tags and self-supervised contrastive learning without negative samples, which can reinforce the ability of our model to excavate stronger discriminative features, especially for occlusion. Extensive experimental results on several benchmark datasets show our proposed model obtains superior Re-ID performance consistently and outperforms the state-of-the-art ReID methods by large margins on the mean average accuracy (mAP) and Rank-1 accuracy.
\end{abstract}


%
\IEEEpeerreviewmaketitle

\section{Introduction}
%
%
%
%
\IEEEPARstart{P}{erson} re-identification (ReID) has been being a fundamental yet challenging computer vision task, which aims at associating the person of the same identity across multiple non-overlapping cameras \cite{author1}. In past decade, with the fast development of deep learning technology, person ReID has made great progress and been dominated by the deep learning-based methods for a long time \cite{author2,author3,author4}. However, person ReID task confronts some complex challenges, such as occlusion, pose changes, view variation and background clutter, so it is the fucus of current research to how to extract robust and discriminate features to combat them.

At present, most existing person ReID methods \cite{author49} assume that the entire human body is visible in camera views which hence can't generalize well to camera scenes with a lot of occlusion, because the occluded camera scenes may be contain plenty of incomplete body information. Since humans are occluded often occurs in natural scenes, occluded person ReID \cite{author10,author13,author18} has attracted much attention from many researchers recently. By reviewing the occluded person ReID methods \cite{author5}, we find that occluded person ReID mainly solve the following problems: 1) the occlusion will introduce the noisy information, which leads to mismatch; 2) the occlusion may have the similar features to human body parts, resulting in failing to learn more discriminative features; 3) the human pose changes, camera view variation and inter-frame human movement may cause feature misalignment. However, lots of ReID methods \cite{author6,author7} adopted the common data augmentation strategies (e.g., color distortion, random horizontal flipping and random erasing) to enhance the feature representation, which is only suited to the normal scenes not existing the occlusion and fails to deal with the occluded person images because these models lack the occlusion data for their training. 

Recently, Vision Transformer (ViT) \cite{author9} adopts the information of all image patches to extract the global features by self-attention mechanism, which has been proven to have good performance and outperform the CNN-based approaches \cite{author52,author53} in image classification task. However, the ability of ViT to extract local features is weak because the difference between image patch embeddings will decrease gradually and the spatial information contained in them will became blurred with the processing of self-attention blocks. This weakness of ViT has little impact on general visual tasks such as image classification \cite{author9} and face recognition \cite{author48} but is susceptible to interference for the occlusion scenes \cite{author10}. Thus, in order to solve this problem, different from the strategy of occlusion suppression in \cite{author10}, inspirited by the paper \cite{author8},  we proposed a novel data augmentation strategy with random rectangle mask, which can improve the performance of ViT model under occlusion conditions.
 
In addition, as an efficient representation learning method, contrastive learning shows great potential in person ReID, especially for self-supervised and unsupervised learning, which makes the model learn more distinguishable feature representations so as to enhance the recognition performance by contrasting the feature differences between different samples. For unsupervised contrastive learning, how to generate the pseudo labels and update them is the key step, which usually relies on some clustering algorithms to gradually approximate the true label distribution. Unlike it, self-supervised contrastive learning realizes the representation learning without manual annotation by some specially designed tasks such as image transformation prediction and image context prediction \cite{author51}, which opens up a new way for person ReID. Furthermore, there are also some methods \cite{author10,author68} based on self-supervised contrastive learning for the occlusion problem, and they usually solve the occlusion problem by a per-trained encoder or predictor. Although these methods have made significant progress, it has come at a high training cost. Thus, in this paper, we proposed a novel joint-training framework combining self-supervised and supervised learning, which aims at combating the occlusion challenge in a more cost-effective way. The main contributions of this work are summarized as follows:
\begin{itemize}
  \item We designed a novel random rectangle mask strategy to simulate the occlusion in real scenes so as to learn more robust feature representation for occlusion. In order to enhance the ability of feature representation further, we still cleverly fused other image enhancing methods such as Gaussian blur \cite{author15}, Random color jittering and Solarization \cite{author50}.
  \item We builded a self-supervised contrastive learning branch based on the sharing ViT construction, which doesn't rely on the negative samples sampling in traditional contrastive learning and directly adopts the inherent structure information of the image to enhance the feature learning ability of the ViT encoder by self-supervision.
  \item We utilized the joint-training loss function to integrate the advantages of supervised learning with ID tags and self-supervised contrastive learning without negative samples, which can promote the performance of our model.
  \item Quantitative evaluation and ablation study on several authoritative benchmark datasets show that the proposed model and its each components are reasonable and effective.
\end{itemize}

 The remaining of this work is organized as follows: Section \ref{sec:rw} introduces the related works. In section \ref{sec:PM}, we elaborate the proposed transformer-based person re-identification framework and data augmentation method designed for its contrastive learning branch. Section \ref{sec:exp} shows the ablation study and evaluation results on several benchmark person Re-ID datasets. Finally, conclusions are explained in Section \ref{sec:con}.
\section{Related works}\label{sec:rw}
\subsection{Occluded person re-identification}
The occluded person re-identification mainly confronts the challenges of incomplete body information and spatial misalignment. The existing occluded person ReID solutions can be classified into two categories, methods based on the feature alignment of visible region cues \cite{author19,author20,author21} and methods based on information completion \cite{author34,author23,author24}.

The core idea of methods based on the feature alignment of visible region cues is to accurately locate the visible regions of person in the image, which only relies on the semantic information of visible region to align the feature representations of different body parts and  match the pedestrians with the same identity in other camera view. Miao \emph{et al}. \cite{author19} proposed a feature alignment method based on the human key-points information, taking advantage of human key-points information to guide the model to focus on the non-occluded regions and only these are utilized for the retrieval. He \emph{et al.} \cite{author20} designed a deep spatial feature reconstruction method for partial person ReID, which realized the implicit feature alignment when calculating the reconstruction error of the spatial feature maps. However, solving the reconstruction error needs the time-consuming computation and is not suitable for the large scale re-identification task. Chen \emph{et al.} \cite{author21} proposed an occlusion-aware mask network, which utilizes the attention-guided mask module to precisely capture body parts regardless of the occlusion. Wang \emph{et al.} \cite{author22} proposed a novel feature erasing and diffusion network to simultaneously handle the interference from non-pedestrian occlusions and non-target pedestrians, improving the model's perception ability towards target pedestrians and robustness towards non-target pedestrians.

The person ReID methods based on information completion usually adopt the spatiotemporal context to recover the pedestrian information of the occluded parts so as to improve the features' discrimination \cite{author34,author24}. Xu \emph{et al.} \cite{author23} proposed a feature recovery transformer to exploit the pedestrian information in its $k$-nearest neighbors features for occluded feature recovery. In \cite{author24}, a spatiotemporal feature completion module was proposed to recover the occluded body parts via a region-encoder and a region-decoder for video person ReID. 

Different from these methods above, we presented a joint-training model combining the supervised and self-supervised learning, which only introduces our specially designed data augmentation method in the training stage to learn more robust feature representation for the occlusion, not increasing the inference cost.
\subsection{Transformer-based person re-identification}
When capturing the long range dependencies of features, the CNN-based methods confront some challenges due to the limitation of receptive fields. In contrast, the ViT can overcome it by the self-attention mechanism, demonstrating excellent performance in some areas such as image understanding, image segmentation and visual question and answer. TransReID \cite{author4}, as the first person ReID model to use the pure transformer architecture for extracting features, has further promoted the development of this field. The PFT \cite{author26} utilized the proposed three modules (patch full dimension enhancement module, fusion and reconstruction module and spatial slicing module) to enhance the efficiency of vision transformer. The PFD \cite{author27} utilized pose information to clearly disentangle semantic components (e.g. human body or joint parts) and selectively match non-occluded parts correspondingly, improving the performance of person ReID effectively. Li \emph{et al.} \cite{author28} proposed a novel end-to-end part-aware transformer for occluded person ReID via a transformer encoder-decoder architecture, including a pixel context based transformer encoder and a part prototype based transformer decoder. In \cite{author10}, the authors proposed the occlusion suppression and repairing transformer, including a self-supervised occlusion predictor, occlusion suppression encoder and feature repairing head, which can enhance the model's ability of extracting discriminative local features for the occlusion scenes. There are other some methods to try to use the transformer to aggregate the features extracted from the CNN backbone, e.g., GiT \cite{author29}, HaT \cite{author30} and so on.
\subsection{Self-supervised learning}
Recently, self-supervised learning is widely applied to person re-identification task because it can mine the useful information from unlabelled data and has become an important direction of person ReID research, which can be classified into contrastive learning and masked image modeling.

Contrastive learning is aimed at learning similar/dissimilar representations from sample pairs with same semantic and ones with different semantic. The traditional contrastive learning usually relies on the annotated datasets where each image has a clear identity label. The core idea of this model is to attract the positive sample pairs and repulse the negative sample pairs by a reasonable designed loss function (e.g., Triplet loss \cite{author54}, InfoNCE \cite{author55}, contrastive loss \cite{author31} and so on) so as to learn the more discriminative features. Different from it, the self-supervised contrastive learning constructs positive and negative sample pairs from the strong and normal enhancing results of the same image, which can learn the meaningful features representation from lots of unlabelled data by the pre-designed tasks in the absence of labelled data. The existing self-supervised contrastive learning methods mainly include MoCo \cite{author56}, SimCLR \cite{author15}, BYOL \cite{author17}, SimSiam \cite{author37} and so on.

Masked image modeling was first proposed in \cite{author39}, which can enhance the ability of representation learning by performing pretext tasks such as image/feature reconstruction. Its basic idea is to randomly mask some image patches and reconstruct them from visible image patches by ViT in a self-supervised manner. This technology not only can be transferred to downstream computer vision tasks such as image classification and instance segmentation, but also provided a new way to solve the occluded person re-identification. In this paper, inspirited by the masked image modeling, we designed a novel data augmentation method for contrastive learning branch, that is random rectangle mask strategy.
\section{Methodology}\label{sec:PM}
In this section, we first describes the data augmentation method designed for our contrastive learning branch in subsection \ref{sec:da}. Next, in subsection \ref{sec:dba}, we introduce our proposed transformer-based person re-identification framework in details. Finally, we give the joint-training loss combining supervised and self-supervised learning in the proposed model in subsection \ref{sec:jointloss}.
\subsection{Data augmentation}\label{sec:da}
\begin{figure} 
  \centering
  \includegraphics[width=8cm]{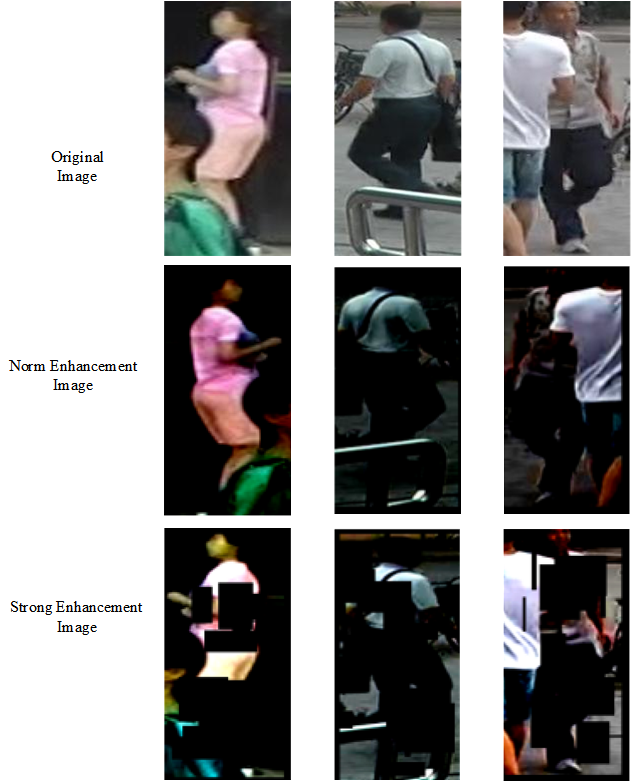}\\
  \caption{Visualization of the different data augmentation methods.}\label{fig_vis}
\end{figure}

Empirically, the occlusion often occurs for the person re-identification tasks. The normal data augmentation methods, e.g., random cropping, flipping and random color distortion, are not very effective in the face of occlusion. Thus, in order to improve the model's performance under the situation of partial occlusion and incomplete information, inspirited by the paper \cite{author8}, we designed a novel data augmentation method to imitate the occlusion in real scenes, that is random rectangle mask strategy, whose implementing method is given in detail in Algorithm \ref{Alg:1} and the visualized results see the figure \ref{fig_vis}. This data augmentation method can help the model learn more discriminative features so as to facilitate effective recognition and classification in the case of incomplete information. In addition, in order to further enhance the model's generalization, we also use other data augmentation methods, such as Gaussian blur, random color jittering and Solarization. 
\begin{algorithm}
\hspace*{0.02in}\textbf{Input:} The image: ${\rm I}$; the masked ratio: $r$; the maximum mask size: $M=(m_{h},m_{w})$.\\
\textbf{Output:} the masked image: ${\rm I}_{mask}$\\
\vspace{-12pt}
\begin{algorithmic}[1]
\STATE Initialize $h,w\leftarrow{\rm I.shape}$, $mask\leftarrow zeros(h,w)$, $area_{total}\leftarrow h\ast w$, $area_{mask}\leftarrow aera_{total}\ast r$, $area_{current}\leftarrow 0$\\
/*Calculate the target masked area*/
\WHILE{$area_{current}<area_{mask}$}
{
\STATE  $rect_{height},rect_{width}\leftarrow random(1,M)$\\
\STATE $x_{start}\leftarrow random(0,h-rect_{height})$
\STATE $y_{start}\leftarrow  random(0,w-rect_{width})$
\STATE $mask_{temp}\leftarrow zeros(h,w)$
\STATE $mask_{temp}(x_{start}:x_{start}+rect_{height},y_{start}:y_{start}+rect_{width})\leftarrow true$
\STATE $mask_{com}\leftarrow mask_{temp}\cup mask$
\STATE $aera_{new}\leftarrow sum(mask_{com})$
 \IF {$aera_{new}<=aera_{mask}$}
   \STATE $mask\leftarrow mask_{com}$
   \STATE $aera_{current}\leftarrow aera_{new}$\\
   /*Adjust $rect_{height}, rect_{width}$ to ensure the final area doesn't exceed $area_{mask}$ */
 \ELSE
     \STATE $aera_{remain}\leftarrow aera_{mask}-aera_{current}$
     \STATE $factor \leftarrow sqrt(area_{remain} / (rect_{height}\ast rect_{width}))$
     \vspace{-10pt}
     \STATE $rect_{height}, rect_{width}\leftarrow (rect_{height},rect_{width})\ast factor$
     \STATE $mask_{temp}\leftarrow zeros(h,w)$
     \STATE $mask_{temp}(x_{start}:x_{start}+rect_{height},y_{start}:y_{start}+rect_{width})\leftarrow true$
     \STATE $mask\leftarrow mask_{temp}\cup mask$
     \STATE $aera_{current}\leftarrow sum(mask)$
 \ENDIF
}
\ENDWHILE
\RETURN ${\rm I}_{mask}\leftarrow {\rm I}\odot mask$ /*$\odot$ denotes the element-wise product.*/
\end{algorithmic}
\caption{Random rectangle mask strategy}
\label{Alg:1}
\end{algorithm}
\subsection{Dual branch architecture}\label{sec:dba}
\begin{figure*} 
  \centering
  \includegraphics[width=16cm]{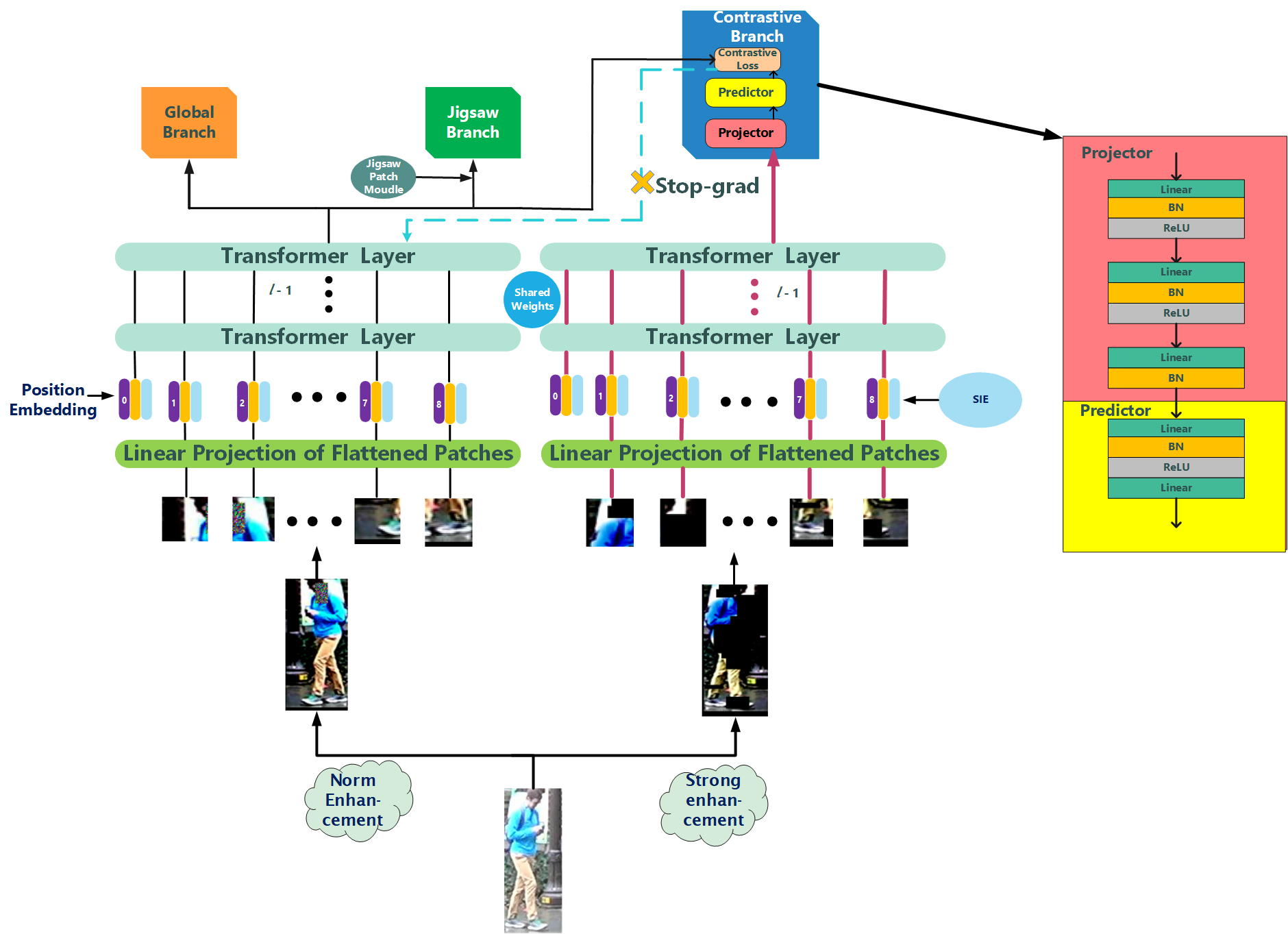}\\
  \caption{The overall framework of self-supervised and supervised combining transformer-based person re-identification.}\label{fig_reid}
\end{figure*}

In \cite{author4}, the TransReID has been proven to a well-performing person re-identification model based on the transformer. Based on it, we build our self-supervised and supervised combining transformer-based person re-identification framework, shown in Figure \ref{fig_reid}. Our model consists of the normal augmentation and strong augmentation branches, and they utilize the TransReID as the feature extractor and share the same weight of transformer layers. Specifically, each image is first processed by inputting the normal augmentation and strong augmentation branches respectively, where the normal augmentation includes some common methods, such as random horizontal flipping, random cropping, random erasing and so on, and the strong augmentation adopts the data augmentation methods mentioned in subsection \ref{sec:da}. Next, each augmented image is split into $N$ fixed-sized patches, and each patch is mapped to a $D$ dimensions vector by a linear projection. Subsequently, their patch embedding, position embedding and side information embedding are fed into $l-1$ layers transformer encoder together. Finally, the output features of the normal augmentation branch are fed into the global predicted head and jigsaw predicted head to further extract the features and calculate the ID loss and triplet loss, and the output features of the strong augmentation branch are input the projector and predictor in sequence to further process, obtaining features and the output ones of the normal branch calculate the similarity by the contrastive learning together. Our proposed model not only retains the Side Information Embedding (SIE) and jigsaw patch modules of the TransReID, but also assists features learning in the training phase by the contrastive learning branch, which greatly improves the model's performance in complex scenes without affecting the efficiency of inference, especially for the occlusion.
\subsection{Joint-training loss combining self-supervised and supervised learning}\label{sec:jointloss}
Similar to the TransReID, we optimize the normal augmentation branch network by building ID loss and triplet loss for global features and $K$ fine-grained local features that obtained by the jigsaw patch module. The ID loss $\mathcal{L}_{ID}$ is the cross-entropy loss without label smoothing. For a triplet set $\{a,p,n\}$, the triplet loss $\mathcal{L}_{T}$ with soft-margin is defined as following
\begin{equation}\label{eq1}
 \mathcal{L}_{T}=\log[1+\exp(\|f_{a}-f_{p}\|_{2}^{2}-\|f_{a}-f_{n}\|_{2}^{2})].
\end{equation}
Thus, for the normal augmentation branch, the total loss of supervised learning can be represented as
\begin{equation}\label{eq2}
 \mathcal{L}_{TS}=\mathcal{L}_{ID}(f_{g})+\mathcal{L}_{T}(f_{g})+\frac{1}{K}\sum_{k=1}^{K}(\mathcal{L}_{ID}(f_{l}^{k})+\mathcal{L}_{T}(f_{l}^{k})),
\end{equation}
where $f_{g}$ is served as the global feature obtained by the standard transformer encoder, and $f_{l}^{k}$ denotes the output token of the $k$-th group obtained by re-grouping them after shuffling the patches through the $l$-th transformer encoder.

In order to enhance the representation ability of the transformer encoder, we assist the training of ReID task by self-supervised contrastive learning. In the contrastive learning branch, we adopt the same loss function as the paper \cite{author37}, which can be defined as following:
\begin{equation}\label{eq3}
  \mathcal{L}_{contrast}=\frac{1}{2}\mathcal{D}(p_{1},stopgrad(z_{2}))+\frac{1}{2}\mathcal{D}(p_{2},stopgrad(z_{1})),
\end{equation}
where $p_{1},p_{2}$ respectively denote the output vectors by the predictor of two stream branches, $z_{1},z_{2}$ denote the output features of the $l-1$-th transformer encoder of two stream architectures, $stopgrad(\cdot)$ is the stop-gradient operation which means that $z_{1}, z_{2}$ are treated as the constant, that is, only receiving gradients from $p$ and no gradient from $z$ when encoding the corresponding input image, and $\mathcal{D}(\cdot,\cdot)$ is the negative cosine similarity.

Thus, the final total loss is defined as following:
\begin{equation}\label{eq4}
  \mathcal{L}=\lambda\ast\mathcal{L}_{TS}+(1-\lambda)\ast\mathcal{L}_{contrast},
\end{equation}
where $\lambda$ is the hyper-parameter for balancing two losses. by adjusting it reasonably, we can let the contrastive loss assist the training of ReID task efficiently so as to utilize the advantage of self-supervised learning to enhance the performance and robustness of our proposed model against complex and variable scenes. 
\section{Experiments}\label{sec:exp}
In this section, we first introduce the benchmark datasets and evaluation metrics used in our experiments. Next, we describe the implementation details. Then, we compare our proposed method with the state-of-the-art person Re-ID methods on several benchmark person Re-ID datasets. Finally, we conduct ablation study to analyze the effect of weights setting of patch linear projection layer, data augmentation methods and hyper-parameter $\lambda$ in our proposed person Re-ID model. 
\subsection{Datasets and evaluation metric}
\noindent\textbf{Datasets.} To verify the performance of our proposed model, we compare it with other some the state-of-the-art person ReID methods on five person ReID benchmark datasets, including Market-1501 \cite{author40}, MSMT17 \cite{author42}, DukeMTMC-reID \cite{author41}, Occluded-Duke \cite{author19} and Occluded-Market \cite{author57}. The details of these datasets are summarized in Table \ref{tab:data}, where Occluded-Duke is formed by applying a special division on DukeMTMC-reID and Occluded-Market is formed by integrating and re-partitioning MARS \cite{author58} and Market-1501 \cite{author40}, but the occlusion ratio in the training set of Occluded-Market reaches 63\%, which is much higher than that in Occluded-Duke.\\

\begin{table*}
  \caption{The scale of five different person ReID datasets.$\gamma$ is the occlusion rario (\%).}
  \centering
  \begin{tabular}{ccccccccccc}
    \toprule
     \multirow{2}{*}{Dataset}& \multirow{2}{*}{\#cam}\ & \multicolumn{3}{c}{Training dataset} & \multicolumn{3}{c}{Query} &\multicolumn{3}{c}{Gallery}\\
     \cmidrule(r){3-5} \cmidrule(r){6-8} \cmidrule(r){9-11}
           &   &ID &Image &$\gamma$ &ID &Image &$\gamma$ &ID &Image &$\gamma$ \\
    \midrule
    Market-1501&6&751&12936&-&750&19732&-&750&3368&-\\
    MSMT17&15&1041&32621&-&3060&11659&-&3060&82161&-\\
    DukeMTMC-reID&8&702&16522&-&702&2228&-&702&17661&-\\
    Occluded-Duke&8&702&15618&9&702&2210&100&702&17661&10\\
    Occluded-Market&-&780&9287&63&533&2343&100&751&15913&8\\
    \bottomrule
  \end{tabular}
  \label{tab:data}
\end{table*}

\noindent\textbf{Evaluation metric.} For the fairness of comparison, we follow conventions in the ReID community and adopt the cumulative matching characteristic (CMC) curve and the mean average precision (mAP) as evaluation metrics. Specifically, we adopt the first element of the CMC curve, i.e., Rank-1, which indicates the hit rate of the first match.
\subsection{Implementation details}
In this paper, all the experiments of our proposed method are run on a single NVIDIA A100 GPU with 40G RAM. All input person images are resize to $256\times128$. For the normal augmentation branch, the training images are augmented with random horizontal flipping, random cropping and random erasing, and for another branch, they are augmented by the methods mentioned in subsection \ref{sec:da}. The maximum mask size $M$ is set to $(128,128)$ in Algorithm \ref{Alg:1}. The batch size is set to 100 with 4 images per ID. Stochastic Gradient Descent (SGD) optimizer is employed with a momentum of 0.9 and the weight decay of $1e-4$. The base learning rate is set to $0.0125$ with cosine learning rate decay. The projection head and prediction head are respectively a 3-layer MLP and 2-layer MLP, the hidden layers of which are 768-d and 4096-d and are with ReLU; the output layers are both 256-d, without ReLU. The hyper-parameter $\lambda$ is set to 0.95. The parameters of patch projecting layer in our model are randomly initialized and frozen rather than the learnable ones in the traditional methods. The initial weights of ViT adopt the officially released per-trained model, the model of which is first trained on Image-21K and then fine-tuned on ImageNet-1K.
\subsection{Comparison with state-of-the-art methods} 
\noindent\textbf{Comparison on Occluded Person ReID Datasets.} On Occluded-Duke and Occluded Market, we compare our proposed model with other state-of-the-art methods which is classified into two categories: CNN-based ReID methods and Transformer-based ReID method, the results of which are shown in Table \ref{tab:occ-result}. We find that our proposed method obtains the competitive performance on the Occluded-Duke dataset. Specifically, comparing with the most related method, e.g., TransReID, our proposed model obtains the 1.8\% and 2.8\% performance gains in terms of the mAP and Rank-1 scores. It's noting that our proposed model does not surpass some ReID methods that used the feature recovery (e.g., FRT \cite{author23} and OSRTrans \cite{author10}) or extra clues (PFD \cite{author27}), but these methods either utilize the pre-trained occlusion predictor to predict the occluded regions and recover them by some designed reconstructing model (e.g., GAN or ViT) or use the extra clues (e.g, pose information \cite{author27}, which is only usually obtained by the trained pose estimation model) to selectively match non-occluded parts, which needs other extra models and increases the inference time. Compared to them, our proposed model does not require any pre-trained module or negative samples but achieves comparable performance. With the increase of occluded samples in the training set, our proposed model obtains the highest mAP and Rank-1 scores (e.g., Occluded Market, the occlusion ratio of its training set rises to 63\%, seeing the Table \ref{tab:data}), which is superior to the state-of-the-art OSRTrans \cite{author10} by 4.2\% and 1.3\% in terms of mAP and Rank-1. In addition, we also notice that our proposed model exceeds the second place TransReID \cite{author4} by 2.0\% in mAP score. These experimental results show that our proposed model is more adaptive to different datasets, unlike the feature recover is susceptible to domain gap.
\begin{table*}
  \caption{Performance comparison with state-of-the-art methods on Occluded-Duke and Occluded-Market.Extra-clue denotes using external model. ${*}$means the encoder is with a small step sliding-window setting. Bold values indicate the best performance, and underlined values indicate the second-best performance.}
  \centering
  \begin{tabular}{cccccc}
    \toprule
     \multirow{2}{*}{Method}& \multirow{2}{*}{Extra-clue} & \multicolumn{2}{c}{Occluded-Duke} & \multicolumn{2}{c}{Occluded-Market} \\
     \cmidrule(r){3-4} \cmidrule(r){5-6}
           &   &mAP &Rank-1 &mAP &Rank-1 \\
    \midrule
    \multicolumn{6}{c}{Transformer-Based} \\
    \midrule
    TransReID$^{*}$\cite{author4}&\XSolidBrush&59.2&66.4&\underline{69.7}&80.2\\
    FED\cite{author22}&\XSolidBrush&56.4&68.1&53.3&66.7\\
    PFD\cite{author27}&\CheckmarkBold&\textbf{61.8}&69.5&-&-\\
    PFT\cite{author26}&\XSolidBrush&60.8&69.8&-&-\\
    FRT\cite{author23}&\CheckmarkBold&61.3&\underline{70.7}&-&-\\
    PAT\cite{author28}&\XSolidBrush&53.6&64.5&-&-\\
    DRL-Net\cite{author46}&\XSolidBrush&53.9&65.8&-&-\\
    OSRTrans\cite{author10}&\XSolidBrush&\underline{61.5}&\textbf{72.9}&67.5&\underline{82.0}\\
    \midrule
    \multicolumn{6}{c}{CNN-Based} \\
    \midrule
    HOReID\cite{author59}&\CheckmarkBold&43.8&55.1&49.3&64.9\\
    MSOSNet\cite{author45}&\XSolidBrush&56.3&68.6&-&-\\
    QPM\cite{author60}&\XSolidBrush&49.7&64.4&-&-\\
    PVPM\cite{author61}&\CheckmarkBold&42.9&54.2&49.4&66.8\\
    MVI$^{2}$P\cite{author43}&\XSolidBrush&57.3&68.6&-&-\\
    SCSRL\cite{author57}&\XSolidBrush&51.4&62.6&54.5&73.8\\
    AET-Net\cite{author47}&\XSolidBrush&54.5&64.5&-&-\\
    \midrule
    Ours$^{*}$&\XSolidBrush&61.0&69.2&\textbf{71.7}&\textbf{83.3}\\
    \bottomrule
  \end{tabular}
  \label{tab:occ-result}
\end{table*}

\noindent\textbf{Comparison on General Person ReID Datasets.} We also compare our proposed model with some state-of-the-art methods on three general person ReID datasets, including Market1501, MSMT17 and DukeMTMC-ReID, the results of which are shown in Table \ref{tab:result}. Seeing from it, we find that our SSSC-TransReID ranks No.1 among these person ReID methods on Market1501 and MSMT17 datasets, especially obtaining the improvement of 1.5\% and 0.4\% than the second place TransReID \cite{author4} in terms of mAP and Rank-1 scores on MSMT17. In addition, it is worth noting that our proposed method outperforms the state-of-the-art OSRTrans that used the occlusion suppression and repairing, respectively rising 1.1\%/0.5\%  and 3.3\%/1.7\%in terms of mAP/Rank-1 on Market1501 and DukeMTMC-ReID datasets, which illustrates that our proposed method have better performance than the ReID method that specifically designed for the occluded ReID dataset on the general ReID dataset. The main reason is that the occluded ReID models are designed to fit the occluded scenarios and focus on dealing with occlusion interference, so they can not extract features from each detail region of the image, which affects their ability to generalize across domains. But our method has good adaptability to different domains and is not sensitive to domain gap. Although the performance of our method is slightly inferior to that of PFD \cite{author27}, different from the PFD, our method does not need pre-training human pose estimation model to extract pose information to guide the local features aggregation so as to selectively match non-occluded parts. Therefore, our proposed model is simpler and more efficient in inference. In a word, our SSSC-TransReID not only performs well and has strong robustness on Occluded ReID datasets, but also achieve excellent performance on the general ReID datasets.  
\begin{table*}
  \caption{Performance comparison with state-of-the-art methods on Market1501 , MSMT17 and DukeMTMC-reID datasets.Extra-clue denotes using external model.${*}$means the encoder is with a small step sliding-window setting.Bold values indicate the best performance, and underlined values indicate the second-best performance.}
  \centering
  \begin{tabular}{cccccccc}
    \toprule
     \multirow{2}{*}{Method}& \multirow{2}{*}{Extra-clue} & \multicolumn{2}{c}{Market1501} & \multicolumn{2}{c}{MSMT17}  & \multicolumn{2}{c}{DukeMTMC-reID} \\
     \cmidrule(r){3-4} \cmidrule(r){5-6} \cmidrule(r){7-8}
           &   &mAP &Rank-1 &mAP &Rank-1 &mAP &Rank-1 \\
    \midrule
    \multicolumn{8}{c}{Transformer-Based} \\
    \midrule
    TransReID$^{*}$\cite{author4}&\XSolidBrush&88.9&95.2&\underline{67.4}&\underline{85.3}&82.0&90.7\\
    FED\cite{author22}&\XSolidBrush&86.3&95.0&-&-&78.0&89.4\\
    PFD\cite{author27}&\CheckmarkBold&\underline{89.7}&95.5&-&-&\textbf{83.2}&\underline{91.2}\\
    PFT\cite{author26}&\XSolidBrush&88.8&95.3&-&-&82.1&90.7\\
    FRT\cite{author23}&\CheckmarkBold&88.1&95.5&-&-&81.7&90.5\\
    PAT\cite{author28}&\XSolidBrush&88.0&95.4&-&-&78.2&88.8\\
    DRL-Net\cite{author46}&\XSolidBrush&86.9&94.7&-&-&76.6&88.1\\
    OSRTrans\cite{author10}&\XSolidBrush&88.7&95.3&-&-&79.6&89.9\\
    \midrule
    \multicolumn{8}{c}{CNN-Based} \\
    \midrule
    OSNet\cite{author62}&\XSolidBrush&84.9&94.8&52.9&78.7&73.5&88.6\\
    MSOSNet\cite{author45}&\XSolidBrush&88.8&\underline{95.7}&60.1&82.3&-&-\\
    MVI$^{2}$P\cite{author43}&\XSolidBrush&87.9&95.3&61.4&83.9&-&-\\
    SCSRL\cite{author57}&\XSolidBrush&86.2&95.0&53.2&78.9&-&-\\
    AET-Net\cite{author47}&\XSolidBrush&87.5&94.8&-&-&80.1&89.5\\
    \midrule
    Ours$^{*}$&\XSolidBrush&\textbf{89.8}&\textbf{95.8}&\textbf{68.9}&\textbf{85.7}&\underline{82.9}&\textbf{91.6}\\
    \bottomrule
  \end{tabular}
  \label{tab:result}
\end{table*}
\subsection{Visualize and analysis}
\begin{figure} 
  \centering
  \includegraphics[width=7cm]{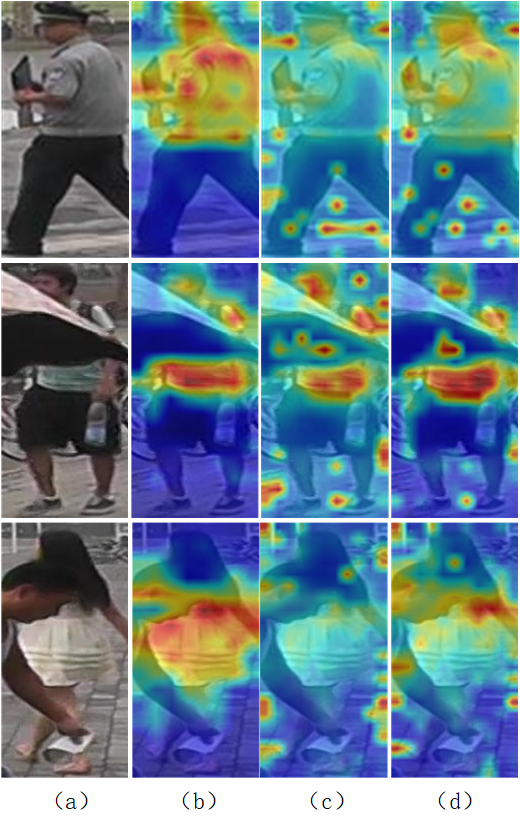}\\
  \caption{Grad-CAM visualization of attention maps. (a) Input images, (b) TransReID without using sliding window, (c) TransReID with sliding window, (d) our proposed method}\label{fig_heat}
\end{figure}
In order to clearly clarify why our proposed method perform well on the occluded and general ReID datasets,  we use Grad-CAM \cite{author63} to visualize the attention maps of the encoders from TransReID and our method by the heatmap, shown in Figure \ref{fig_heat}, which selects two occluded and one unoccluded images. Seeing from the Figure \ref{fig_heat}, comparing with the TransReID, our proposed method not only keeps paying attention to the overall features of target but also is not sensitive to the occlusion. For the occluded person, our proposed method makes the encoding features more focus on the unoccluded parts of person, which efficiently reduces the effect of the occlusion. This property of our proposed method ensures that it can extract more robust and discriminative features for person ReID task on the occluded and unoccluded images, which further verifies the superiority of our proposed method in person ReID task.
\subsection{Ablation study}
On Occluded-Duke dataset, we perform a number of ablation experiments to validate the effectiveness of our proposed method from different aspects.
\begin{table}[!h]
  \caption{Impact of different setting methods of Patch Linear Projection Layer weights.}
  \centering
  \begin{tabular}{ccc}
    \toprule
    \multirow{1}{*}{Method}& \multirow{1}{*}{mAP} & \multicolumn{1}{c}{Rank-1}\\
    \midrule
    Learned&60.3&68.5\\
    Frozen&\textbf{61.0}&\textbf{69.2}\\
    \bottomrule
  \end{tabular}
  \label{tab:weight}
\end{table}

\noindent\textbf{Weights setting of patch linear projection layer.} Previous studies \cite{author38} have found that the use of ViT as backbone will affect the performance and stability of the model. In order to explore the effect of different setting methods of patch linear projection layer weights for the model performance, we compare two different patch linear projection layer weights, one is the learned weights, that is linear projection layer weights are updated with the training processing after they are randomly initialized; another is the frozen weights, namely linear projection layer weights are not updated when our model is trained. Table \ref{tab:weight} shows the mAP and Rank-1 scores of our proposed method under two different setting of patch linear projection layer weights. Seeing from it, the frozen weights method is superior to the learned one by 0.7\% and 0.7\% in terms of mAP and Rank-1 scores, which illustrates that freezing patch linear project layer weights after random initialization can make our proposed model's training more stable and perform better.

\begin{figure} 
  \centering
  \includegraphics[width=8cm]{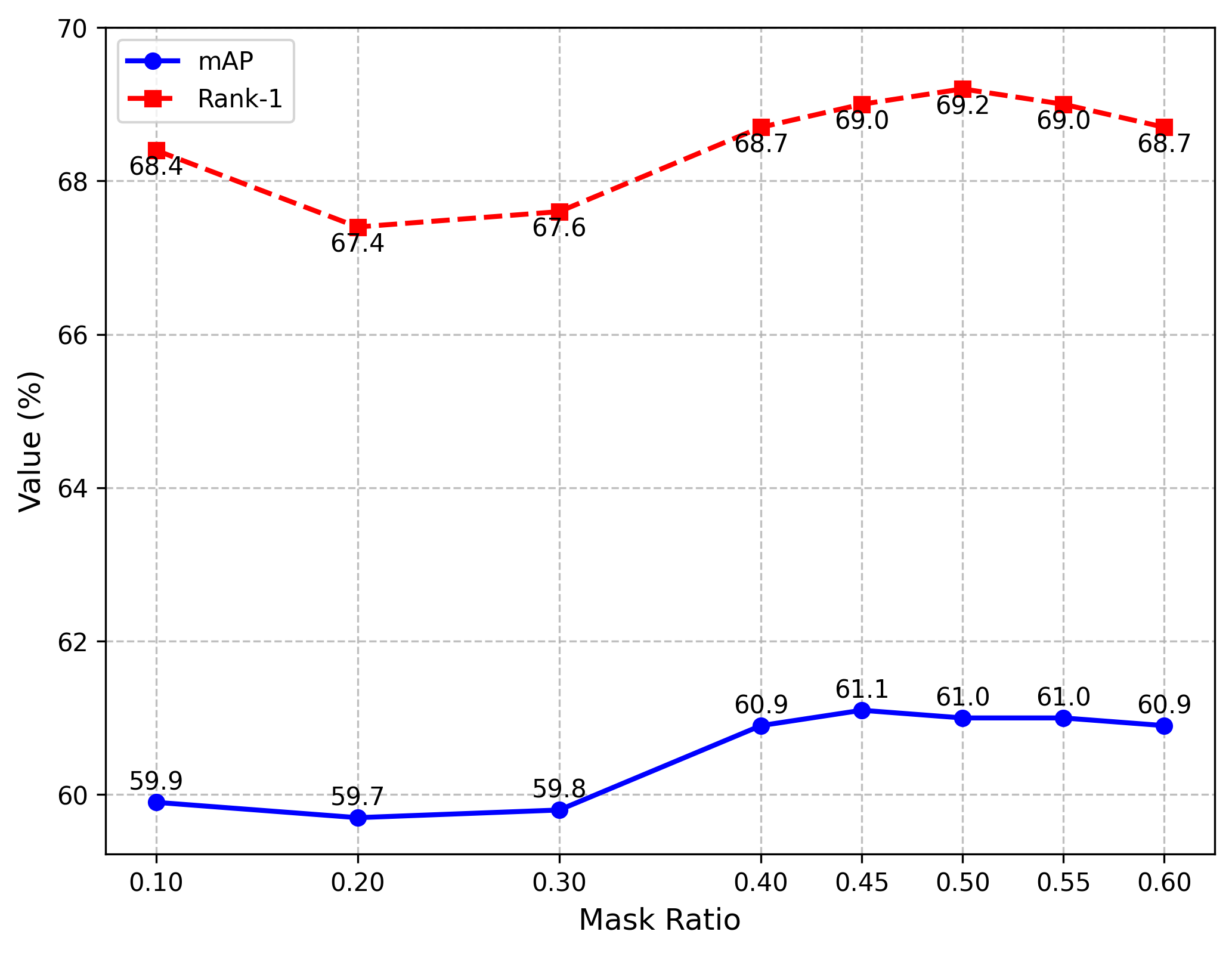}\\
  \caption{Analysis of the impact of different mask ratios for our data augmentation method proposed in subsection \ref{sec:da}.}\label{fig_weight}
\end{figure}

\noindent\textbf{Data augmentation.} We analysed the impacts of different mask ratios and data augmentation combining methods for our model performance. In Figure \ref{fig_weight}, we plot the curves of mAP and Rank-1 scores when selected different mask ratios in our proposed data augmentation method. Seeing from it, the mAP and Rank-1 scores of our proposed method fluctuate with respect to mask ratio when it varies from 0.1 to 0.6 with a step of 0.1. In order to obtain the best mask ratio, the step size is further reduced to 0.05 when approaching the estimated peak. When the mask ratios are set to 0.45 and 0.5, the mAP and Rank-1 scores of our method respectively reach the highest points. As the mask ratio continues to increase, the performance of our method becomes worse. For the person ReID task, the Rank-1 metric is more important than the mAP one, and when the mask ratio is set to 0.45 or 0.5, there is only a slight gap of 0.1 for the mAP scores of our method. Consequently, in our experiments, the mask ratio of our proposed data augmentation method is set to 0.5, namely the mask ratio that the Rank-1 score reaches the peak.

For strong augmentation branch, we compare our random mask with four most similar data augmentation methods, the results of which are shown in Table \ref{tab:aug_com}. Although these data augmentation methods are designed to simulate occlusion so as to improve the generalization ability of the model, we observe that our random mask and Hide-and-seek perform better than other three methods. The main reason is that our random mask and Hide-and-seek simulate more severe occlusion, compared with other three  methods (the occlusion ratio of each sample is set to  0.5 for our random mask and Hide-and-seek, and the occlusion ratio of each sample is randomly selected from 0 to 0.5 for other three methods), which is beneficial for the contrastive learning. Seeing from the Table \ref{tab:aug_com}, our random mask is superior to Hide-and-seek. Even though they have the same mAP score,  the Rank-1 score of our random mask is 0.6\% higher than the Rank-1 one of Hide-and-seek. The possible reason is that the occlusion simulated by our method is more diverse and complex, and the one generated by Hide-and-seek is uniformly and randomly distributed, because more complex and diverse samples are helpful for contrastive learning. In Table \ref{tab:data_aug}, we compare the results of our random mask in combination with other several data augmentation methods (Color Jitter, Gaussian Blur and Solarize), and we find that our Rank-1 score rises by 0.5\% when combining simultaneously with those three augmentation methods.
\begin{table}
  \caption{ Comparison of different data augmentation methods on the strong augmentation branch.}
  \centering
  \begin{tabular}{c|cccc}
    \toprule
    \multirow{1}{*}{Augmention}& \multicolumn{1}{c}{mAP}& \multicolumn{1}{c}{Rank-1}& \multicolumn{1}{c}{Rank-5}& \multicolumn{1}{c}{Rank-10}\\
    \midrule
     Random Erasing\cite{author67}&59.7&67.3 &81.8 &86.2  \\
     CutOut\cite{author66}&60.7&67.6&82.4&\textbf{87.6} \\
     CutMix\cite{author65}&60.8&68.1&82.7&87.3 \\
     Hide-and-seek\cite{author64}&\textbf{61.0}&68.1&82.5&87.2 \\
     Random mask (ours)&\textbf{61.0}&\textbf{68.7}&\textbf{83.0}&\textbf{87.6} \\
    \bottomrule
  \end{tabular}
  \label{tab:aug_com}
\end{table} 
\begin{table*}
  \caption{Impact of different data augmentation combining methods.}
  \centering
  \begin{tabular}{cccccc}
    \toprule
    \multirow{1}{*}{Random Mask (ours)} & \multicolumn{1}{c}{Color Jitter}& \multicolumn{1}{c}{Gaussian Blur}& \multicolumn{1}{c}{Solarize}& \multicolumn{1}{c}{mAP}& \multicolumn{1}{c}{Rank-1}\\
    \midrule
      \CheckmarkBold & & & &61.0&68.7 \\
      \CheckmarkBold &\CheckmarkBold & & &60.5&68.1 \\
      \CheckmarkBold & & \CheckmarkBold& &60.1&68.0 \\
      \CheckmarkBold & & & \CheckmarkBold&60.9&68.3 \\
      \CheckmarkBold & \CheckmarkBold&\CheckmarkBold & &60.9&68.8 \\
      \CheckmarkBold & & \CheckmarkBold&\CheckmarkBold &60.7&68.0 \\
      \CheckmarkBold & \CheckmarkBold& &\CheckmarkBold &61.0&68.3 \\
     \CheckmarkBold &\CheckmarkBold &\CheckmarkBold &\CheckmarkBold &\textbf{61.0}&\textbf{69.2} \\
    \bottomrule
  \end{tabular}
  \label{tab:data_aug}
\end{table*}

\begin{figure} 
  \centering
  \includegraphics[width=8cm]{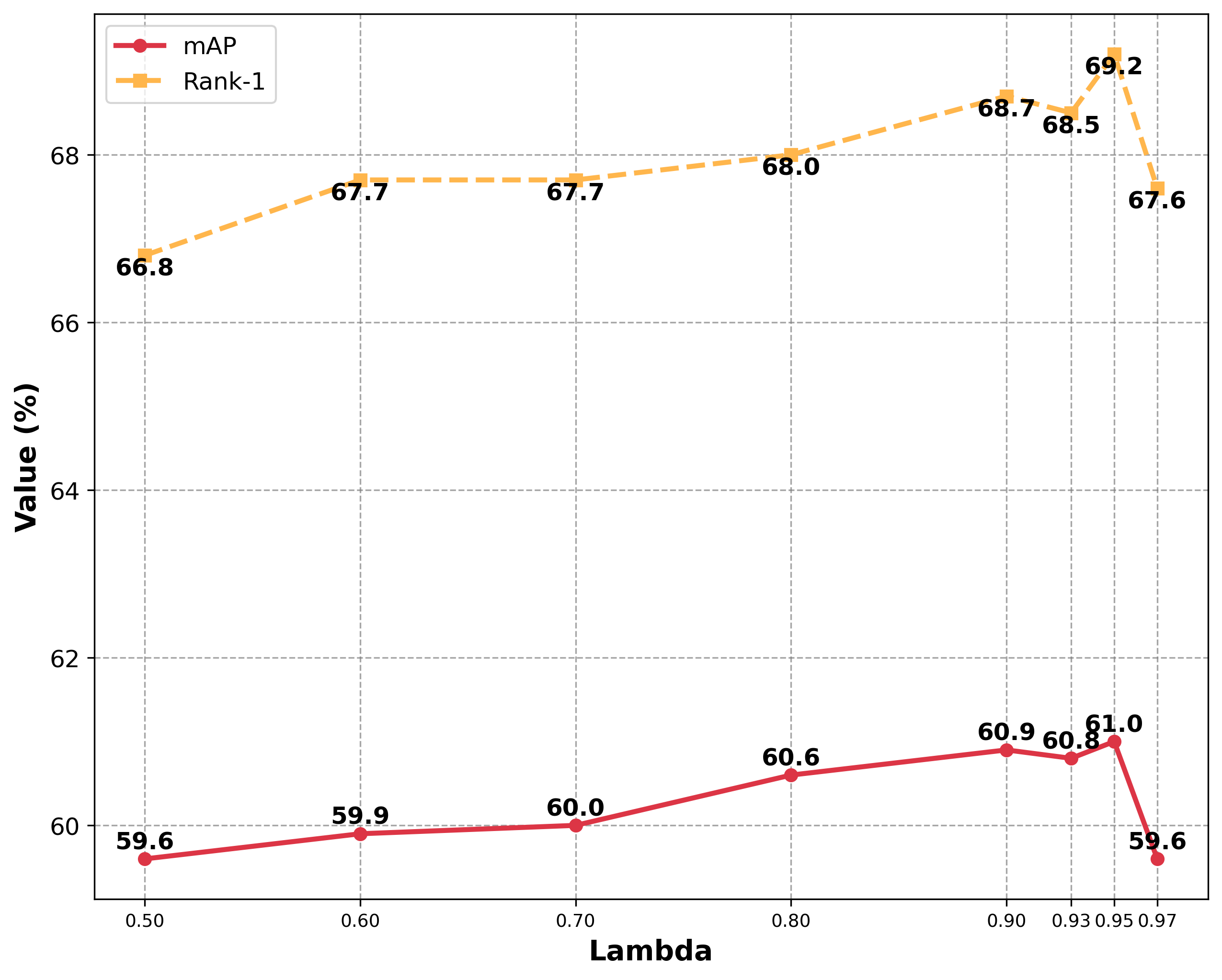}\\
  \caption{Analysis of the impact of the hyper-parameter $\lambda$ of the total loss function.}\label{fig_parameter}
\end{figure}

\noindent\textbf{Effect of hyper-parameter $\lambda$.} The hyper-parameter $\lambda$ is used to balance the supervised learning loss and self-supervised contrastive learning loss of our model. In Figure \ref{fig_parameter}, we plot the curves of mAP and Rank-1 scores when the hyper-parameter $\lambda$ is set to 0.5 and increases in some certain steps $\{0.1, 0.1,0.1,0.1,0.03,0.02,0.02\}$ until 0.97 in our final loss function. We observe that our proposed method obtains the best mAP and Rank-1 scores when the hyper-parameter $\lambda=0.95$, which illustrates that the labeled data dominates in our model training. If $\lambda$ is too small or too large, the performance of our proposed method degrades significantly. When $\lambda$ is decreased to 0.9, the mAP and Rank-1 scores of our method respectively reduce by 0.1\% and 0.5\%. If $\lambda$ continues to decrease slowly, the performance of our method become worse. However, when $\lambda$ increases to 0.97, the performance of our method degrades severely, the mAP and Rank-1 scores of which respectively drop by 1.4\% and 1.6\%. The main reason is that the larger $\lambda$ makes the self-supervised contrastive learning no longer work. Thus, selecting an appropriate hyper-parameter $\lambda$ is extremely important for robust person ReID. In our experiments, the  hyper-parameter $\lambda$ in the equation \ref{eq4} is set to 0.95.

\section{Conclusion}\label{sec:con}
In this paper, we propose a novel self-supervision and supervision combining transformer-based Re-ID framework, which can excavate stronger discriminative features by integrating advantages of supervised learning with ID tags and self-supervised contrastive learning without negative samples. In order to train our model, we design a novel data augmentation method based on random rectangle mask for the contrastive learning branch. The experimental results on several benchmark  datasets demonstrate that our proposed model is superior to the state-of-the-art person ReID methods on the mean average accuracy (mAP) and Rank-1 accuracy. Based on it, we believe that integrating self-supervised and supervised learning has great potential to be further explored for ReID tasks.


%

%

\section*{Acknowledgment}
This work is supported by the National Natural Science Foundation of China (No.61602288) and Fundamental Research Program of Shanxi Province (No.20210302123443). The authors also would like to thank the anonymous reviewers for their valuable suggestions.

\ifCLASSOPTIONcaptionsoff
  \newpage
\fi



%
{\small
\bibliographystyle{IEEEtran}
\bibliography{referenceBib}
}
\end{document}